# Markov Determinantal Point Processes

**Raja Hafiz Affandi**
Department of Statistics
The Wharton School
University of Pennsylvania
rajara@wharton.upenn.edu

**Alex Kulesza**
Dept. of Computer and Information Science
University of Pennsylvania
kulesza@cis.upenn.edu

**Emily B. Fox**
Department of Statistics
The Wharton School
University of Pennsylvania
ebfox@wharton.upenn.edu

## Abstract

A determinantal point process (DPP) is a random process useful for modeling the combinatorial problem of subset selection. In particular, DPPs encourage a random subset $Y$ to contain a *diverse* set of items selected from a base set $\mathcal{Y}$. For example, we might use a DPP to display a set of news headlines that are relevant to a user's interests while covering a variety of topics. Suppose, however, that we are asked to sequentially select *multiple* diverse sets of items, for example, displaying new headlines day-by-day. We might want these sets to be diverse not just individually but also through time, offering headlines today that are unlike the ones shown yesterday. In this paper, we construct a *Markov* DPP (M-DPP) that models a *sequence* of random sets $\{Y_t\}$. The proposed M-DPP defines a stationary process that maintains DPP margins. Crucially, the induced union process $Z_t \equiv Y_t \cup Y_{t-1}$ is also marginally DPP-distributed. Jointly, these properties imply that the sequence of random sets are encouraged to be diverse both at a given time step as well as across time steps. We describe an exact, efficient sampling procedure, and a method for incrementally learning a quality measure over items in the base set $\mathcal{Y}$ based on external preferences. We apply the M-DPP to the task of sequentially displaying diverse and relevant news articles to a user with topic preferences.

## 1 INTRODUCTION

Consider the combinatorial problem of subset selection. Binary Markov random fields are commonly applied in this setting, and in the case of positive correlations, yield subsets that favor similar items. However, in many applications there is naturally a sense of *repulsion*. For example, repulsive processes arise in nature—trees tend to grow in the least occupied space (Neeff et al., 2005), and ant hill locations are likewise over-dispersed relative to uniform placement (Bernstein and Gobbel, 1979). Likewise, many practical tasks can be posed in terms of diverse subset selection. For example, one might want to select a set of frames from a movie that are representative of its content. Clearly, diversity is preferable to avoid redundancy; likewise, each frame should be of high quality. A motivating example we consider throughout the paper is the task of selecting a diverse yet relevant set of news headlines to display to a user. One could imagine employing binary Markov random fields with negative correlations, but such models often involve notoriously intractable inference problems.

Determinantal point processes (DPPs), which arise in random matrix theory (Mehta and Gaudin, 1960; Ginibre, 1965) and quantum physics (Macchi, 1975), are a class of *repulsive processes* and are natural models for subset selection problems where diversity is preferred. DPPs define the probability of a subset in terms of the determinant of a kernel submatrix, and with an appropriate definition of the kernel matrix they can be interpreted as inherently balancing quality and diversity. DPPs are appealing in practice since they offer interpretability and tractable algorithms for exact inference. For example, one can compute marginal and conditional probabilities and perform exact sampling. DPPs have recently been employed for human pose estimation, search diversification, and document summarization (Kulesza and Taskar, 2010, 2011a,b).

In this paper, our focus is instead on modeling diverse *sequences* of subsets. For example, in displaying news headlines from day to day, one aims to select articles that are relevant and diverse on any given day. Additionally, it is desirable to select articles that are diverse relative to those previously shown. We construct a *Markov* DPP (M-DPP) for a sequence of random sets $\{Y_t\}$. The proposed M-DPP defines a stationary process that maintains DPP margins, implying that $Y_t$ is encouraged to be diverse at time $t$. Crucially, the induced union process $Z_t \equiv Y_t \cup Y_{t-1}$ is also marginally DPP-distributed. Since this property implies the diversity of $Z_t$, in addition to the individual diversity of $Y_t$ and $Y_{t-1}$, we conclude that $Y_t$ is diverse from $Y_{t-1}$.

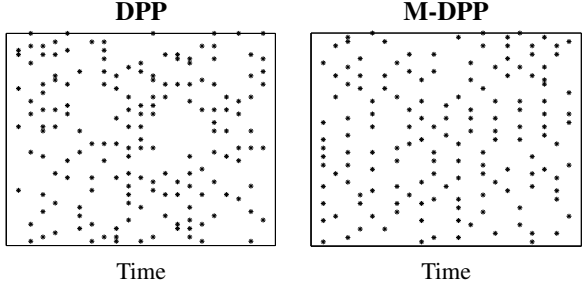

Figure 1: A set of points on a line ($y$ axis) drawn from a DPP independently over time (left) and from a M-DPP (right). While DPP points are diverse only within time steps (columns), M-DPP points are also diverse across time steps.

For an illustration of the improved overall diversity when sampling from a M-DPP rather than independent sequential sampling from a DPP, see Fig. 1.

Our specific construction of the M-DPP yields an exact sampling procedure that can be performed in polynomial time. Additionally, we explore a method for incrementally learning the quality of each item in the base set $\mathcal{Y}$ based on externally provided preferences. In particular, a decomposition of the DPP kernel matrix has an interpretation as defining the quality of each item and pairwise similarities between items. Our incremental learning procedure assumes a well-defined similarity metric and aims to learn features of items that a user deems as preferable. These features are used to define the quality scores for each item. The M-DPP aids in the exploration of items of interest to the user by providing sequentially diverse results.

We study the empirical behavior of the M-DPP on a news task where the goal is to display diverse and high quality articles. Compared to choosing articles based on quality alone or to sampling from an independent DPP at each time step, we show that the M-DPP produces articles that are significantly more diverse across time steps without large sacrifices in quality. Furthermore, within a time step the M-DPP chooses articles with diversity comparable to that of the independent DPP; this is a direct result of the fact that the M-DPP maintains DPP margins. We also consider learning the quality function over time from the feedback of a user with topic preferences. In this setting, the M-DPP returns high quality results that are preferred by the user while simultaneously exploring the topic space more quickly than baseline methods, leading to improved coverage.

## 2 DETERMINANTAL POINT PROCESSES

A random point process $\mathcal{P}$ on a discrete base set $\mathcal{Y} = \{1, \ldots, N\}$ is a probability measure on the set $2^{\mathcal{Y}}$ of all subsets of $\mathcal{Y}$. Let $K$ be a semidefinite matrix with rows and columns indexed by the elements of $\mathcal{Y}$. $\mathcal{P}$ is called a determinantal point process (DPP) if there exists $K \preceq I$ (all eigenvalues less than or equal to 1) such that if $\boldsymbol{Y}$ is a random set drawn according to $\mathcal{P}$, then for every $A \subseteq \mathcal{Y}$:

$$\mathcal{P}(\boldsymbol{Y} \supseteq A) = \det(K_A) . \quad (1)$$

Here, $K_A \equiv [K_A]_{i,j \in A}$ denotes the submatrix of $K$ indexed by elements in $A$, and we adopt the convention that $\det(K_\emptyset) = 1$. We will refer to $K$ as the marginal kernel. If we think of $K_{ij}$ as measuring the similarity between items $i$ and $j$, then

$$\mathcal{P}(\boldsymbol{Y} \supseteq \{i, j\}) = K_{ii}K_{jj} - K_{ij}^2 \quad (2)$$

implies that $\boldsymbol{Y}$ is unlikely to contain both $i$ and $j$ when they are very similar; that is, a DPP can be seen as modeling a collection of diverse items from the base set $\mathcal{Y}$.

DPPs can alternatively be constructed via L-ensembles (Borodin and Rains, 2005). An L-ensemble is a probability measure on $2^{\mathcal{Y}}$ defined via a positive semidefinite matrix $L$ indexed by elements of $\mathcal{Y}$:

$$\mathcal{P}_L(\boldsymbol{Y} = A) = \frac{\det(L_A)}{\det(L + I)} , \quad (3)$$

where $I$ is the $N \times N$ identity matrix. It can be shown that an L-ensemble is a DPP with marginal kernel $K = L(I+L)^{-1}$. Conversely, a DPP with marginal kernel $K$ has L-ensemble kernel $L = K(I - K)^{-1}$ (when the inverse exists).

An intuitive way to think of the L-ensemble kernel $L$ is as a Gram matrix (Kulesza and Taskar, 2010):

$$L_{ij} = q_i \phi_i^\top \phi_j q_j , \quad (4)$$

interpreting $q_i \in \mathbb{R}^+$ as representing the intrinsic *quality* of an item $i$, and $\phi_i, \phi_j \in \mathbb{R}^n$ as unit length feature vectors representing the *similarity* between items $i$ and $j$ with $\phi_i^\top \phi_j \in [-1, 1]$. Under this framework, we can model quality and similarity separately to encourage the DPP to choose *high quality* items that are *dissimilar* to each other. This is very useful in many applications. For example, in response to a search query we can provide a very relevant (i.e. high quality) but diverse (i.e. dissimilar) list of results.

**Conditional DPPs** For any $A, B \subseteq \mathcal{Y}$ with $A \cap B = \emptyset$, it is straightforward to show that

$$\mathcal{P}_L(\boldsymbol{Y} = A \cup B | \boldsymbol{Y} \supseteq A) = \frac{\det(L_{A \cup B})}{\det(L + I_{\mathcal{Y} \setminus A})} , \quad (5)$$

where $I_{\mathcal{Y} \setminus A}$ is a matrix with ones on the diagonal entries indexed by the elements of $\mathcal{Y} \setminus A$ and zeros elsewhere.

This conditional distribution is itself a DPP over the elements of $\mathcal{Y} \setminus A$ (Borodin and Rains, 2005). In particular, suppose $\boldsymbol{Y}$ is DPP-distributed with L-ensemble kernel $L$, and condition on the fact that $\boldsymbol{Y} \supseteq A$. Then the set $\boldsymbol{Y} \setminus A$ is DPP-distributed with marginal and L-ensemble kernels

$$K^A = \left[ I - (L + I_{\mathcal{Y} \setminus A})^{-1} \right]_{\mathcal{Y} \setminus A} \quad (6)$$

$$L^A = \left( \left[ (L + I_{\mathcal{Y} \setminus A})^{-1} \right]_{\mathcal{Y} \setminus A} \right)^{-1} - I . \quad (7)$$

Here, $[\cdot]_{\mathcal{Y}\setminus A}$ denotes the submatrix of the argument indexed by elements in $\mathcal{Y}\setminus A$. Thus, DPPs as a class are closed under most natural conditioning operations.

In selecting a diverse collection of elements in $\mathcal{Y}$, a DPP jointly models both the size of a set and its content. In some applications, the goal is to select (diverse) sets of a fixed size. In order to achieve this goal, we can instead consider a fixed-size determinantal point processes, or $k$DPP (Kulesza and Taskar, 2011a), which gives a distribution over all random subsets $Y \subseteq \mathcal{Y}$ with fixed cardinality $k$. The L-ensemble construction of a $k$DPP, denoted $\mathcal{P}_L^k$, gives probabilities

$$\mathcal{P}_L^k(\boldsymbol{Y} = A) = \frac{\det(L_A)}{\sum_{|B|=k}\det(L_B)} \qquad (8)$$

for all sets $A$ with cardinality $k$ and any positive semidefinite kernel $L$. Kulesza and Taskar (2011a) developed efficient algorithms to normalize, sample and marginalize $k$DPPs using properties of elementary symmetric polynomials.

## 3 MARKOV DETERMINANTAL POINT PROCESSES (M-DPPS)

In certain applications, such as in the task of displaying news headlines, our goal is not only to generate a diverse collection of items at one time point, but also to generate collections of items at subsequent time points that are both highly relevant and dissimilar to the previous collection. To address these goals, we introduce the Markov determinantal point process (M-DPP), which emphasizes both marginal and conditional diversity of selected items. Harnessing the *quality* and *similarity* interpretation of the DPP in (4), the M-DPP provides a dynamic way of selecting high quality and diverse collections of items as a temporal process.

We constructively define a first-order, discrete-time autoregressive point process on $\mathcal{Y}$ by specifying a Markov transition distribution (and initial distribution). Throughout, we use the notation $\{\boldsymbol{Y}_t\}$, $\boldsymbol{Y}_t \subseteq \mathcal{Y}$, to represent a sequence of sets following a M-DPP. We consider two such constructions: one based on marginal kernels, and the other on L-ensembles. Both yield equivalent stationary processes with DPP margins. Additionally, and quite intuitively, the induced union process $\{\boldsymbol{Z}_t \equiv \boldsymbol{Y}_t \cup \boldsymbol{Y}_{t-1}\}$ has DPP margins with a closely related kernel. Combining these two properties, we conclude that the constructed M-DPPs yield a sequence of sets $\{\boldsymbol{Y}_t\}$ that are diverse at any time $t$ and across time steps $t, t-1$.

**Marginal construction.** Define $\mathcal{P}(\boldsymbol{Y}_1 \supseteq A) = \det(K_A)$ and

$$\mathcal{P}(\boldsymbol{Y}_t \supseteq B | \boldsymbol{Y}_{t-1} \supseteq A) = \frac{\det(K_{A\cup B})}{\det(K_A)}, \qquad (9)$$

where $K \prec \frac{1}{2}I$ and $A \cap B = \emptyset$. Throughout, we adopt the implicit constraint that $\boldsymbol{Y}_t \cap \boldsymbol{Y}_{t-1} = \emptyset$. We have immediately the joint probability

$$\mathcal{P}(\boldsymbol{Y}_2 \supseteq B, \boldsymbol{Y}_1 \supseteq A) = \det(K_{A\cup B}), \qquad (10)$$

and therefore

$$\mathcal{P}(\boldsymbol{Y}_2 \supseteq B) = \mathcal{P}(\boldsymbol{Y}_2 \supseteq B, \boldsymbol{Y}_1 \supseteq \emptyset) = \det(K_B). \qquad (11)$$

Inductively, the process is stationary and marginally DPP.

Finally, we have the union of consecutive sets:

$$\begin{aligned}\mathcal{P}(\boldsymbol{Z}_t &\equiv \boldsymbol{Y}_t \cup \boldsymbol{Y}_{t-1} \supseteq C) \\ &= \sum_{A\subseteq C} \mathcal{P}(\boldsymbol{Y}_t \supseteq C\setminus A, \boldsymbol{Y}_{t-1} \supseteq A) = \sum_{A\subseteq C}\det(K_C) \\ &= 2^{|C|}\det(K_C) = \det((2K)_C). \qquad (12)\end{aligned}$$

That is, $\boldsymbol{Z}_t$ is marginally distributed as a DPP with marginal kernel $2K$. Since a randomly sampled subset of a DPP-distributed set also follows a DPP, marginally we can imagine this process as sampling $\boldsymbol{Z}_t$ and then splitting its elements randomly into two sets, $\boldsymbol{Y}_{t-1}$ and $\boldsymbol{Y}_t$.

**L-ensemble construction.** The above is appealingly simple, but the marginal form of the conditional in (9) is not particularly conducive to a sequential sampling process. Instead, we can rewrite everything as L-ensembles. Assume that at the first time step $\mathcal{P}(\boldsymbol{Y}_1 = Y_1) = \frac{\det(L_{Y_1})}{\det(L+I)}$ and define the transition distribution as

$$\mathcal{P}(\boldsymbol{Y}_t = Y_t | \boldsymbol{Y}_{t-1} = Y_{t-1}) = \frac{\det(M_{Y_t \cup Y_{t-1}})}{\det(M + I_{\mathcal{Y}\setminus Y_{t-1}})}, \qquad (13)$$

for $M = L(I-L)^{-1}$. Note that the transition distribution is essentially a conditional DPP with L-ensemble kernel $M$ (Eq. (5)). $M$ is well-defined as long as $L \prec I$, which is equivalent to $K \prec \frac{1}{2}I$, as in the marginal construction.

Now we have the joint probability

$$\mathcal{P}(\boldsymbol{Y}_2 = Y_2, \boldsymbol{Y}_1 = Y_1) = \frac{\det(M_{Y_1 \cup Y_2})}{\det(M + I_{\mathcal{Y}\setminus Y_1})} \frac{\det(L_{Y_1})}{\det(L+I)}. \qquad (14)$$

Using the fact that $\det(M + I_{\mathcal{Y}\setminus Y_1})/\det(M+I) = \det(L_{Y_1})$,

$$\mathcal{P}(\boldsymbol{Y}_2 = Y_2, \boldsymbol{Y}_1 = Y_1) = \frac{1}{\det(L+I)} \frac{\det(M_{Y_1 \cup Y_2})}{\det(M+I)}. \qquad (15)$$

Therefore, marginally,

$$\begin{aligned}\mathcal{P}(\boldsymbol{Y}_2 = Y_2) &= \sum_{Y_1 \subseteq \mathcal{Y}} \frac{1}{\det(L+I)} \frac{\det(M_{Y_1 \cup Y_2})}{\det(M+I)} \\ &= \sum_{(Y_1 \cup Y_2) \supseteq Y_2} \frac{1}{\det(L+I)} \frac{\det(M_{Y_1 \cup Y_2})}{\det(M+I)} \\ &= \frac{\det(M + I_{\mathcal{Y}\setminus Y_2})}{\det(L+I)\det(M+I)} = \frac{\det(L_{Y_2})}{\det(L+I)}.\end{aligned} \qquad (16)$$

Here, we used $\sum_{B \supseteq A} \det(M_B) = \det(M + I_{\mathcal{Y} \setminus A})$, which is immediately derived from (5). By induction, we conclude

$$\mathcal{P}(\boldsymbol{Y}_t = Y_t) = \frac{\det(L_{Y_t})}{\det(L + I)} . \qquad (17)$$

Thus, our construction yields a stationary process with $\boldsymbol{Y}_t$ marginally distributed as a DPP with L-ensemble kernel $L$.

One can likewise analyze the margin of the induced union process $\{\boldsymbol{Z}_t \equiv \boldsymbol{Y}_t \cup \boldsymbol{Y}_{t-1}\}$:

$$\mathcal{P}(\boldsymbol{Z}_t \equiv \boldsymbol{Y}_t \cup \boldsymbol{Y}_{t-1} = C)$$
$$= \sum_{A \subseteq C} \mathcal{P}(\boldsymbol{Y}_t = C \setminus A, \boldsymbol{Y}_{t-1} = A)$$
$$= \sum_{A \subseteq C} \frac{1}{\det(L+I)} \frac{\det(M_C)}{\det(M+I)}$$
$$= \frac{2^{|C|}}{\det(L+I)} \frac{\det(M_C)}{\det(M+I)} \qquad (18)$$
$$= \frac{1}{\det(L+I)} \frac{\det((2M)_C)}{\det(M+I)} . \qquad (19)$$

Noting that

$$\det(M+I)\det(L+I) = \det((M+I)(L+I))$$
$$= \det((M+I)(I - (M+I)^{-1} + I))$$
$$= \det(2M + 2I - I) = \det(2M + I) , \qquad (20)$$

we conclude

$$\mathcal{P}(\boldsymbol{Z}_t \equiv \boldsymbol{Y}_t \cup \boldsymbol{Y}_{t-1} = C) = \frac{\det((2M)_C)}{\det(2M+I)} . \qquad (21)$$

We have shown that $\boldsymbol{Z}_t$ is marginally distributed as a DPP with L-ensemble kernel $2M$. The corresponding marginal kernel is

$$2M(2M+I)^{-1} = 2L(I-L)^{-1} \left[(L+I)(I-L)^{-1}\right]^{-1}$$
$$= 2L(L+I)^{-1} = 2K . \qquad (22)$$

Thus, we have reproduced the same characterization of $\boldsymbol{Z}_t$ as in (12) for the marginal kernel construction.

To summarize the marginal properties of the M-DPP, using the notation $\boldsymbol{Y} \sim L, K$ to denote that $\boldsymbol{Y}$ is from a DPP with L-ensemble kernel $L$ and marginal kernel $K$, we have:

$$\boldsymbol{Y}_t \sim L, K \qquad (23)$$
$$\boldsymbol{Z}_t \sim 2L(I-L)^{-1}, 2K . \qquad (24)$$

### 3.1 COMMENTS ON THE M-DPP

While we have shown that the M-DPP subsets are diverse at subsequent time steps, this does not necessarily imply diversity at longer intervals. In fact, it is possible for realizations to have oscillations, where groups of high-quality items recur every two (or more) time steps. However, this is not necessarily a problem in practice for several reasons. First, the M-DPP construction straightfowardly extends to higher order models with longer memory, if desired. Second, it is possible to show that the M-DPP does not *harm* long-term diversity relative to independent sampling from a DPP. If we make a separate copy of each item at each time step, then the M-DPP can be seen as a large DPP on item/time pairs $(i, t)$. Denoting the marginal kernel by $\hat{K}$, the Markov property implies that $\hat{K}_{(i,t)(j,u)}$ is only nonzero when $|t - u| \leq 1$. The probability of item $i$ appearing at any set of time steps, given by the appropriate determinant of $\hat{K}$, can only be reduced by the off-diagonal entries compared to independent DPP samples at each time step. Thus, the M-DPP can only make global repetition less likely. Finally, in our experiments (Sec. 5) we report results that suggest that M-DPP oscillations do not arise in the task we study.

### 3.2 MARKOV $k$DPPS

One can also construct a *Markov kDPP* (M-$k$DPP). Although we define a stationary process, our construction does not yield $\boldsymbol{Y}_t$ marginally $k$DPP. Instead, the M-$k$DPP simply ensures that $\boldsymbol{Z}_t \equiv \boldsymbol{Y}_t \cup \boldsymbol{Y}_{t-1}$ follows a $2k$DPP. Since $\boldsymbol{Z}_t$ is encouraged to be diverse, the subsets $\boldsymbol{Y}_t$ and $\boldsymbol{Y}_{t-1}$ will likewise be diverse despite not following a $k$DPP themselves.

We start by defining the margin and transition distributions:

$$\mathcal{P}(\boldsymbol{Y}_{t-1} = Y_{t-1}) = \frac{\sum_{|A|=k} \det(L_{Y_{t-1} \cup A})}{\binom{2k}{k} \sum_{|B|=2k} \det(L_B)} \qquad (25)$$

$$\mathcal{P}(\boldsymbol{Y}_t = Y_t | \boldsymbol{Y}_{t-1} = Y_{t-1}) = \frac{\det(L_{Y_{t-1} \cup Y_t})}{\sum_{|A|=k} \det(L_{Y_{t-1} \cup A})} , \qquad (26)$$

where $A$ and $Y_t$ are disjoint from $Y_{t-1}$. Then, jointly

$$\mathcal{P}(\boldsymbol{Y}_t = Y_t, \boldsymbol{Y}_{t-1} = Y_{t-1}) = \frac{\det(L_{Y_{t-1} \cup Y_t})}{\binom{2k}{k} \sum_{|B|=2k} \det(L_B)} , \qquad (27)$$

from which we confirm the stationarity of the process:

$$\mathcal{P}(\boldsymbol{Y}_t = Y_t) = \frac{\sum_{|Y_{t-1}|=k} \det(L_{Y_t \cup Y_{t-1}})}{\binom{2k}{k} \sum_{|B|=2k} \det(L_B)} . \qquad (28)$$

The implied union process has margins

$$\mathcal{P}(\boldsymbol{Z}_t \equiv \boldsymbol{Y}_t \cup \boldsymbol{Y}_{t-1} = C)$$
$$= \sum_{A \subseteq C, |A|=k} \mathcal{P}(\boldsymbol{Y}_t = C \setminus A, \boldsymbol{Y}_{t-1} = A)$$
$$= \sum_{A \subseteq C, |A|=k} \frac{\det(L_C)}{\binom{2k}{k} \sum_{|B|=2k} \det(L_B)}$$
$$= \frac{\det(L_C)}{\sum_{|B|=2k} \det(L_B)} , \qquad (29)$$

which is a $2k$DPP with L-ensemble kernel $L$.

**Algorithm 1** Sampling from a DPP
> **Input:** L-ensemble kernel matrix $L$
> $\{(v_n, \lambda_n)\}_{n=1}^N \leftarrow$ eigenvector/value pairs of $L$
> $J \leftarrow \emptyset$
> **for** $n = 1, \ldots, N$ **do**
>> $J \leftarrow J \cup \{n\}$ with prob. $\frac{\lambda_n}{\lambda_n + 1}$
>
> $V \leftarrow \{v_n\}_{n \in J}$
> $Y \leftarrow \emptyset$
> **while** $|V| > 0$ **do**
>> Select $y_i$ from $\mathcal{Y}$ with $\Pr(y_i) = \frac{1}{|V|} \sum_{v \in V} (v^\top e_i)^2$
>> $Y \leftarrow Y \cup \{y_i\}$
>> $V \leftarrow V_\perp$, an orthonormal basis for the subspace of V orthogonal to $e_i$
>
> **Output:** $Y$

### 3.3 SAMPLING FROM M-DPPS AND M-$k$DPPS

In the previous subsections we showed how our constructions of M-(k)DPPs lead to DPP (and DPP-like) marginals for $\{Y_t\}$ and the union process $\{Z_t\}$. These connections to DPPs give us valuable intuition about the diversity induced both within and across time steps. They serve another purpose as well: since DPPs and $k$DPPs can be sampled in polynomial time, we can leverage existing algorithms to efficiently sample from M-DPPs and M-$k$DPPs.

Hough et al. (2006) first described the DPP sampling algorithm shown in Algorithm 1. The first step is to compute an eigendecomposition $L = \sum_{n=1}^N \lambda_n v_n v_n^\top$ of the kernel matrix; from this, a random subset $V$ of the eigenvectors is chosen by using the eigenvalues to bias a sequence of coin flips. The algorithm then proceeds iteratively, on each iteration selecting a new item $y_i$ to add to the sample and then updating $V$ in a manner that de-emphasizes items similar to the one just selected. Note that $e_i$ is the $i$th elementary basis vector whose elements are all zero except for a one in position $i$. Algorithm 1 runs in time $O(N^3 + Nk^3)$, where $N$ is the number of available items and $k$ is the cardinality of the returned sample.

To adapt this algorithm for sampling M-DPPs, we will proceed sequentially, first sampling $Y_1$ from the initial distribution and then repeatedly selecting $Y_t$ from the transition distribution given $Y_{t-1}$. The initial distribution is a DPP with L-ensemble kernel $L$ and can therefore be sampled directly using Algorithm 1. As shown in Sec. 3, the transition distribution (13) is a conditional DPP with L-ensemble kernel $M = L(L - I)^{-1}$; using (7), the L-ensemble kernel for $Y_t$ given $Y_{t-1} = Y_{t-1}$ can be written as

$$L^{(t)} = \left( (M + I_{\mathcal{Y} \setminus Y_{t-1}})^{-1}_{\mathcal{Y} \setminus Y_{t-1}} \right)^{-1} - I. \quad (30)$$

Thus we can sample simply and efficiently from a M-DPP using Algorithm 2. The runtime is $O(TN^3 + TNk_{\max}^3)$, where $k_{\max}$ is the maximum number of items chosen at a single time step. Note that for constant $k_{\max}$ this is the same

**Algorithm 2** Sampling from a Markov DPP
> **Input:** matrix $L$
> $M \leftarrow L(L - I)^{-1}$
> $Y_1 \leftarrow$ DPP-SAMPLE$(L)$
> **for** $t = 2, \ldots, T$ **do**
>> $L^{(t)} \leftarrow \left( (M + I_{\mathcal{Y} \setminus Y_{t-1}})^{-1}_{\mathcal{Y} \setminus Y_{t-1}} \right)^{-1} - I$
>> $Y_t \leftarrow$ DPP-SAMPLE$(L^{(t)})$
>
> **Output:** $\{Y_t\}$

**Algorithm 3** Sampling from a $k$DPP
> **Input:** L-ensemble kernel matrix $L$, size $k$
> $\{(v_n, \lambda_n)\}_{n=1}^N \leftarrow$ eigenvector/value pairs of $L$
> $J \leftarrow \emptyset$
> **for** $n = N, \ldots, 1$ **do**
>> **if** $u \sim U[0,1] < \lambda_n \frac{e_{k-1}^{n-1}}{e_k^n}$ **then**
>>> $J \leftarrow J \cup \{n\}$
>>> $k \leftarrow k - 1$
>>> **if** $k = 0$ **then**
>>>> **break**
>
> {continue with the rest of Algorithm 1}

runtime as a Kalman filter with a state vector of size $N$.

Kulesza and Taskar (2011a) proved that a modification to the first loop in Algorithm 1 allows sampling from a $k$DPP with no change in the asymptotic complexity. The result is Algorithm 3; here $e_k^n$ denotes the elementary symmetric polynomial $e_k^n = \sum_{|J|=k} \prod_{n \in J} \lambda_n$, which can be computed efficiently using recursion.

We can now use Algorithm 3 to perform sequential sampling for a M-$k$DPP. At first glance, the initial distribution (which is not a $k$DPP) seems difficult to sample; however, from Sec. 3 we know that it can be obtained by harnessing the union process form of (29) and first sampling a $2k$DPP with L-ensemble kernel $L$ and then throwing away half of the resulting items at random. Transitionally, we have a conditional $k$DPP whose kernel can be computed as in (30). Algorithm 4 summarizes the M-$k$DPP sampling process, which runs in time $O(TN^3 + TNk^3)$.

**Algorithm 4** Sampling from a Markov $k$DPP
> **Input:** matrix $L$, size $k$
> $Z_1 \leftarrow k$DPP-SAMPLE$(L, 2k)$
> $Y_1 \leftarrow$ random half of $Z_1$
> **for** $t = 2, \ldots, T$ **do**
>> $L^{(t)} \leftarrow \left( (L + I_{\mathcal{Y} \setminus Y_{t-1}})^{-1}_{\mathcal{Y} \setminus Y_{t-1}} \right)^{-1} - I$
>> $Y_t \leftarrow k$DPP-SAMPLE$(L^{(t)}, k)$
>
> **Output:** $\{Y_t\}$

# 4 LEARNING USER PREFERENCES

A broad class of problems suited to M-($k$)DPP modeling are also applications in which we would like to learn preferences from a user over time. Recall the news headlines scenario. Here, the goal is to present articles on a daily basis that are both relevant to the user's interests and also non-redundant. With feedback from a user in the form of click-through behavior, we can attempt to simultaneously learn features of the articles that the user regards as preferable. While the diversity offered by a M-DPP is intrinsically valuable for this task, e.g., to keep the user from getting bored, in the context of learning it also has an important secondary benefit: it promotes exploration of the preference space.

Consider the following simple learning setup. At each time step $t$, the algorithm shows the user a set of $k$ items drawn from some base set $\mathcal{Y}_t$, for instance, articles from the day's news. The user then provides feedback by identifying each shown item as either preferred or not preferred, perhaps by clicking on the preferred ones. The algorithm then incorporates this feedback and proceeds to the next round. The learner has two goals. First, as often as possible at least some of the items shown to the user should be preferred. Second, over the long term, many different items preferred by the user should be shown. In other words, the algorithm should not focus on a small set of preferred items.

Perhaps the most important consideration in this framework is balancing showing articles that the user is known to like (*exploitation*) against showing a variety of articles so as to discover new topics in which the user is also interested (*exploration*). Neither extreme is likely to be successful. However, using the L-ensemble kernel decomposition in (4), a DPP seeks to propose sets of items that are simultaneously high quality and diverse. The M-DPP takes this a step further and encourages diversity from step to step while maintaining DPP margins, exposing the user to an even greater variety of items without significantly sacrificing quality. Thus, we might expect that M-($k$)DPPs can be used to enable fast and successful learning in this setting.

The tradeoff between exploration and exploitation is a fundamental issue for interactive learning, and has received extensive treatment in the literature on multi-armed bandits. However, our setup is relatively unusual for two reasons. First, we show multiple items per time step, sometimes called the *multiple plays* setting (Anantharam et al., 1987). Second, we use feature vectors to describe the items we choose, allowing us to generalize to unseen items (e.g., new articles); this is a special case of *contextual bandits* (Langford and Zhang, 2007). Each of these scenarios has received some attention on its own, but it is only in combination that a notion of diversity becomes relevant, since we have both the need to select multiple items as well as a basis for relating them. This combination has been considered recently by Yue and Guestrin (2011), who showed an algorithm that yields bounded regret under the assumption that the reward function is submodular. Here, on the other hand, our goal is primarily to illustrate the empirical effects on learning when the items shown at each time step are sampled from a M-DPP. To that end, we propose a very simple quality learning algorithm that appears to work well in practice. Whether formal regret guarantees can be established for learning with M-DPPs is an open question for future work.

## 4.1 SETUP

To naturally accommodate user feedback and transfer knowledge across items, we will consider algorithms that learn a log-linear quality model assigning item $i$ the score

$$q_i = \exp(\theta^\top f_i) \,, \qquad (31)$$

where $f_i \in \mathbb{R}^m$ is a feature vector for item $i$ and $\theta \in \mathbb{R}^m$ is the parameter vector to be learned. Learning iterates between two distinct steps: (1) sampling articles according to the current quality scores and (2) using user feedback to revise the quality scores via updates to $\theta$.

Let $\theta^{(t)}$ denote the parameter vector prior to time step $t$ and let $q_i^{(t)}$ denote the corresponding quality scores for the items $i \in \mathcal{Y}_t$. We initialize $\theta^{(1)} = \mathbf{0}$ so that $q_i^{(1)} = 1$ for all $i \in \mathcal{Y}_1$. (At this point we are effectively in a purely exploratory mode.) Denote the items preferred by the user at iteration $t$ by $\{a_i^{(t)}\}_{i=1}^{R_t}$, and the non-preferred items by $\{b_i^{(t)}\}_{i=1}^{S_t}$. Inspired by standard online algorithms, we define the parameter update rule as follows:

$$\theta^{(t+1)} \leftarrow \theta^{(t)} + \eta \left( \frac{1}{R_t} \sum_{i=1}^{R_t} f_{a_i^{(t)}} - \frac{1}{S_t} \sum_{i=1}^{S_t} f_{b_i^{(t)}} \right) \qquad (32)$$

That is, we add to $\theta$ the average features of the preferred items, and subtract from $\theta$ the average features of non-preferred items. This increases the quality of the former and decreases the quality of the latter. $\eta$ is a learning rate hyperparameter. We can then proceed to the next time step, computing the new quality scores $q_i^{(t+1)} = \exp(\theta^{(t+1)\top} f_i)$ for each $i \in \mathcal{Y}_{t+1}$.

The updated quality scores are then used to select subsequent items to be shown to the user. In order to separate the challenges of learning the quality scores, which is not our primary interest, from the benefits of incorporating the M-DPP, we consider five sampling methods:

- **Uniform.** We ignore the quality scores and choose $k$ items uniformly at random without replacement.

- **Weighted.** We draw $k$ items with probabilities proportional to their quality scores without replacement.

- **kDPP.** We sample the set of items from a $k$DPP with L-ensemble kernel $L$ given by the decomposition in

**Algorithm 5** Interactive learning of quality scores
**Input:** learning rate $\eta$
$\theta^{(1)} \leftarrow \mathbf{0}$
**for** $t = 1, 2, \ldots$ **do**
$\quad q_i^{(t)} \leftarrow \exp(\theta^{(t)\top} f_i) \quad \forall i \in \mathcal{Y}_t$
$\quad$ Select items to display given $q_i^{(t)}$
$\quad\quad$ (using one of the methods described in Sec. 4.1)
$\quad$ Receive user feedback $\{a_i^{(t)}\}_{i=1}^{R_t}$ and $\{b_i^{(t)}\}_{i=1}^{S_t}$
$\quad \theta^{(t+1)} \leftarrow \theta^{(t)} + \eta \left( \frac{1}{R_t} \sum_{i=1}^{R_t} f_{a_i^{(t)}} - \frac{1}{S_t} \sum_{i=1}^{S_t} f_{b_i^{(t)}} \right)$

(4), where $\phi$ is fixed in advance and $q_i$ are the current quality scores.

- **kDPP + heuristic (threshold).** We sample the set of items from a kDPP after removing articles whose similarity to the previously selected articles exceeds a predetermined threshold. At threshold $> 1$, the heuristic is equivalent to the kDPP.

- **M-kDPP.** We sample the set of items from the M-kDPP transition distribution given the items selected at the previous time step. The L-ensemble transition kernel is as in (30), with $L$ defined as for the kDPP.

The learning algorithm is summarized in Algorithm 5.

### 4.2 LIKELIHOOD-BASED ALTERNATIVE

Instead of the additive learning rule proposed above, one could instead take advantage of the probabilistic nature of the M-DPP and perform likelihood-based learning, which has associated theoretical guarantees. In particular, based on a sequence of user feedback, we could solve for the penalized DPP maximum likelihood estimate of $q^{(t)} = [q_1^{(t)}, \ldots, q_N^{(t)}]$ as:

$$\arg\max_q \prod_{t=1}^{t} \mathcal{P}_q(\{a_i^{(t)}\} \subseteq \mathbf{Y}_t, \{b_i^{(t)}\} \cap \mathbf{Y}_t = \emptyset) + \lambda ||q||_2, \tag{33}$$

where $\mathcal{P}_q$ is a DPP with L-ensemble kernel defined by quality scores $q$ and $\lambda$ is a regularization parameter. We have

$$\mathcal{P}_q(\{a_i^{(t)}\} \subseteq \mathbf{Y}_t, \{b_i^{(t)}\} \cap \mathbf{Y}_t = \emptyset) =$$
$$\left(1 - \mathcal{P}_q(\{b_i^{(t)}\} \subseteq \mathbf{Y}_t \mid \{a_i^{(t)}\} \subseteq \mathbf{Y}_t)\right)$$
$$\cdot \mathcal{P}_q(\{a_i^{(t)}\} \subseteq \mathbf{Y}_t), \tag{34}$$

which has computable terms in a DPP given the quality scores $q$. The M-DPP has obvious extensions. However, in both cases the objective function is not convex so computations are intensive and only converge to local maxima. Due to its simplicity and good performance in practice (see Sec. 5.2), we use the heuristic algorithm described previously for illustrating the behavior of the M-DPP.

## 5 EXPERIMENTS

We study the performance of the M-kDPP for selecting daily news items from a selection of over 35,000 New York Times newswire articles obtained between January and June of 2005 as part of the Gigaword corpus (Graff and Cieri, 2009). On each day of a given week, we display 10 articles from a base set of the roughly 1400 articles written that week. This process is repeated for each of the 26 weeks in our dataset. The goal is to choose a collection of articles that is high quality but also diverse, both marginally and between time steps.

To examine performance in the absence of confounding issues of quality learning, we first consider a scenario in which the quality scores are fixed. Here, we measure both the diversity and quality of articles chosen each day by the different methods. We then turn to quality learning based on user feedback to examine how the properties of the M-kDPP influence the discovery of a user's preferences.

### 5.1 FIXED QUALITY

**Similarity** To generate similarity features $\phi_i$, we first compute standard normalized tf-idf vectors, where the idf scores are computed across all 26 weeks worth of articles. We then compute the cosine similarity between all pairs of articles. Due to the sparsity of the tf-idf vectors, these similarity scores tend to be quite low, leading to poor diversity if used directly as a kernel matrix. Instead, we let the similarity features be given by binary vectors where the $j$th coordinate of $\phi_i$ is 1 if article $j$ is among the 150 nearest neighbors of article $i$ in that week based on our cosine distance metric, and 0 otherwise.

**Quality** In the fixed scenario, we need a way to assign quality scores to articles. A natural approach is to score articles based on their proximity to the other articles; this way, an article that is close to many others (as measured by cosine similarity) is considered to be of high quality. In this data set, for example, we find that there is a large cluster of articles that talk about *politics* and articles that fall under this topic generally have much higher quality than articles that talk about, say, *food*. To model this, we compute quality scores as $q_i = \exp(\alpha d_i)$, where $d_i$ is the sum of the cosine similarities between article $i$ and all other articles in our collection and $\alpha$ is a hyperparameter that determines the dynamic range. We chose $\alpha = 5$ for our data set, although a range of values gave qualitatively similar results.

For each method, we sample sets of articles on a daily basis for each of the 26 weeks. To measure diversity within a time step, we compute the average cosine similarity between articles chosen on a given day. We then subtract the result from 1 so that larger values correspond to greater diversity. Diversity between time steps is obtained by measuring the average cosine similarity between each article at time $t$ and the single most similar article at time $t+1$ (or $t+2$ for 2-step diversity), and again subtracting the result from 1. We also report the average quality score of the articles chosen across all 182 days. All measures are averaged over 100 random runs; statistical significance is computed by bootstrapping.

Table 1: Average Diversity and Quality of Selected Articles

| Method | Marginal diversity | 1-step diversity | 2-step diversity | Quality |
|---|---|---|---|---|
| M-$k$DPP | 0.899 | 0.849 | 0.843 | 0.654 |
| $k$-DPP | 0.896 | 0.786 | 0.779 | 0.668 |
| $k$-DPP + heuristic (0.4) | 0.904 | 0.849 | 0.804 | 0.651 |
| $k$-DPP + heuristic (0.2) | 0.946 | 0.891 | 0.889 | 0.587 |
| Weighted Rand. | 0.750 | 0.681 | 0.677 | 0.756 |
| Uniform Rand. | 0.975 | 0.949 | 0.947 | 0.457 |

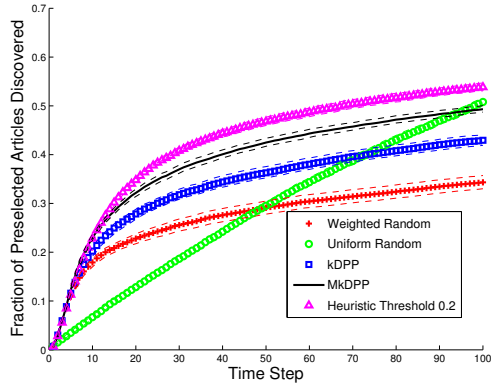

Figure 2: Performance of the methods at recovering the preselected preferred articles. Solid lines indicate the mean over 100 random runs, and dashed lines indicate the corresponding confidence intervals, computed by bootstrapping.

Table 1 displays the results for all methods. The M-$k$DPP shows a marked increase in between-step diversity, on average, compared to the $k$DPP and weighted random sampling. All of the differences are significant at 99% confidence. The average marginal diversities for the M-$k$DPP and $k$DPP are statistically significantly higher than for weighted random sampling, but are not statistically significantly different from each other. This is to be expected since, as we have seen in Sec. 3, the marginal distribution for the M-$k$DPP does not greatly differ from the $k$DPP process. On the other hand, the uniform sampling shows much higher diversity than the other methods, which can be attributed to the fact that it is a purely exploratory method that ignores the quality of the articles it chooses.

Table 1 also shows the average quality of the selected articles. The weighted random sampling chooses, on average, higher quality articles compared to the rest of the methods since it does not have to balance issues of diversity within the set. The $k$DPP on average chooses slightly higher quality articles than the M-$k$DPP, perhaps due to the additional between-step diversity sought by the M-$k$DPP; however, the difference is not statistically significant. It is evident from Table 1 that the M-$k$DPP achieves a balance between the diversity of the articles it chooses (both marginally and across time steps) and their quality.

As for the $k$DPP + heuristic baseline, our experiments show that by tuning the threshold carefully we can mimic the performance of the M-$k$DPP, but without the associated probabilistic interpretation and theoretical properties. When the threshold is too low, quality degrades significantly.

## 5.2 LEARNING PREFERENCES

We also study the performance of the M-$k$DPP when learning from user feedback, as outlined in Sec. 4. For simplicity, we use only a week's worth of news articles (1427 articles). To create feature vectors, we first generate topics by running LDA on the entire corpus (Blei et al., 2003). We then manually label the most prevalent 10 topics as *finance*, *health*, *politics*, *world news*, *baseball*, *football*, *arts*, *technology*, *entertainment*, and *justice*, and associate each article with its LDA-inferred mixture of these topics (a 10-dimensional feature vector $f_i$). We define a synthetic user by a sparse topic preference vector (0.7 for *finance*, 0.2 for *world news*, 0.1 for *politics*, and 0 for all other topics), and preselect as "preferred" the 200 articles whose feature vectors $f_i$ maximize the dot product with the user preference vector.

Similar to our previous experiment, we define the similarity features between articles to be binary vectors based on 50 nearest neighbors using the tf-idf cosine distances. The quality is defined as in Sec. 4, $q_i^{(t)} = \exp(\theta^{(t)\top} f_i)$, where $f_i$ is the feature vector of article $i$ (based on the mixture of topics) normalized to sum to 1. We set the learning rate $\eta = 2$; however, varying $\eta$ did not change the qualitative behavior of each method, only the time scale at which these behaviors became noticeable. We also note that although we base the similarity on 50 nearest neighbors, the results were not sensitive to the size of this neighborhood.

The goal of this experiment is to illustrate how the different methods balance between exploring the space of all articles to discover the 200 preselected articles (*recall*) and exploiting a learned set of features to keep showing preferred articles (*precision*). On one end of the spectrum, uniform sampling simply explores the space of articles without taking advantage of the user feedback, leading to high recall and low precision. On the other end, the weighted random sampling fully exploits the learned preference in selecting articles, but does not have a mechanism to encourage exploration. We demonstrate that the M-$k$DPP balances these two extremes, taking advantage of the user feedback while also exploring diverse articles.

**Results** We use each method to select 10 articles per day over a period of 100 days, using the current quality scores $q^{(t)}$ on each day $t$. We measure recall by keeping track of the fraction of preselected preferred articles (out of the 200 total) that have been displayed so far. We also compute, out of the 10 articles shown on a given day, the fraction that are preferred. This serves as a measure of precision. All measures are averaged over 100 random runs.

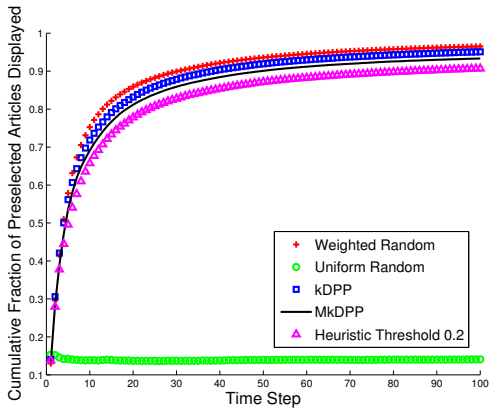

Figure 3: Cumulative fraction of preferred articles displayed to the user.

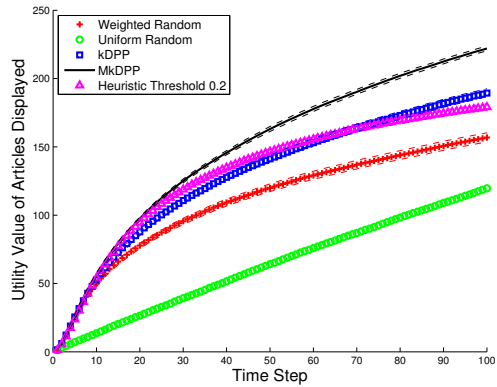

Figure 4: Performance measured by a marginally decreasing utility function.

Figure 2 shows the recall performance of the methods we tested. Uniform sampling discovers the articles at a somewhat linear rate of about 5% per day; given a larger base set relative to the size of the preferred set, however, we would expect a slower rate of discovery. The methods that incorporate user feedback discover a larger set of preferred articles more rapidly by harnessing learned features of the user's interests. The M-$k$DPP dominates both the $k$DPP and weighted random sampling in this metric since it encourages exploration by introducing both marginal and between-step diversity of displayed articles. In contrast, the $k$DPP does not penalize repeating similar marginally diverse sets and the weighted random sampling does not have any explicit mechanism for exploration. It takes uniform random sampling nearly 100 time steps to discover the same number of unique preferred articles as the M-$k$DPP. For the sake of clarity, we omit the results of $k$DPP + heuristic with threshold 0.4 since they are not statistically significantly different from the M-$k$DPP. The supplementary material includes a randomly selected example of the articles displayed on days 99 and 100 for the various methods.

Figure 3 shows the cumulative fraction of displayed articles that were preferred, reflecting precision. (The supplement includes a sample non-cumulative version of Figure 3.) All methods besides uniform sampling quickly achieve high precision. Weighted random sampling displays the largest number of preferred articles per day, almost always having precision of at least 0.9. However, as we have observed, this large precision is at the cost of lower recall. In particular, weighted random sampling quickly homes in on features related to a small subset of preferred articles, thereby increasing the probability of them being repeatedly selected with no force to counteract this behavior. As expected, by only requiring marginal diversity, the $k$DPP achieves slightly higher precision than the M-$k$DPP on average (both typically above 0.8), but again at the cost of reduced explo-ration. Overall, the differences in precision between these methods are not large. In many applications, having 8 out of 10 results preferred may be more than sufficient.

Finally, to examine the balance between exploration and exploitation, we compute a metric based on the idea of marginally decreasing utility. Under this metric, at every time step, the user experiences a utility of 1 for each preferred article shown for the first time. If a previously displayed preferred article is once again chosen, the user gets a utility of $\frac{1}{l+1}$ where $l$ is the number of times that article has appeared in the past. The underlying assumption is that a user benefits from seeing preferred articles, but in decreasing amounts as the same articles are repeatedly displayed. Figure 4 shows the performance of the methods under this utility metric; the M-$k$DPP scores highest.

## 6 CONCLUSION

We introduced the Markov DPP, a combinatorial process for modeling diverse sequences of subsets. By establishing the theoretical properties of this process, such as stationary DPP margins and a DPP union process, we showed how our construction yields sets that are diverse at each time step as well as from one time step to the next, making it appropriate for interactive tasks like news recommendation. Additionally, by explicitly connecting with DPPs, further properties of M-DPPs are straightforwardly derived, such as the marginal and conditional expected set cardinality.

We showed how to efficiently sample from a M-DPP, and found empirically that the model achieves an improved balance between diversity and quality compared to baseline methods. We also studied the effects of the M-DPP on learning, finding significant improvements in recall at minimal cost to precision for a news task with user feedback.

## 7 ACKNOWLEDGMENTS

This work was supported in part by AFOSR Grant FA9550-10-1-0501 and NSF award 0803256.